\documentclass[sigplan,nonacm]{acmart}

\AtBeginDocument{%
  \providecommand\BibTeX{{%
    \normalfont{} B\kern-0.5em{\scshape i\kern-0.25em b}\kern-0.8em\TeX}}
    }
 
\usepackage{fancyhdr}
\usepackage{color}
\usepackage{algorithm}
\usepackage{algorithmic}
\usepackage{eso-pic} % used by \AddToShipoutPicture 
\usepackage{forloop}
\usepackage{tikz} 
\usepackage{amsmath}
\usepackage{inputenc} 
\usepackage{microtype} 
\usepackage{graphicx} 
\usepackage{subcaption}
\usepackage{booktabs}
\usepackage{hyperref}
\usepackage{placeins}
\usepackage{import} 
\graphicspath{{figs/}} %Setting the graphicspath
\usepackage{caption}
\usepackage[newfloat,frozencache=true]{minted}  
\newcommand{\BoGraph}{BoGraph}
\begin{document}

%don't want date printed
\date{}

\title{\BoGraph{}: Structured Bayesian Optimization From Logs for Expensive Systems with Many Parameters}
\author{Sami Alabed}
\orcid{0000-0001-8716-526X}
\affiliation{% 
  \institution{University of Cambridge, UK
    \\ The Alan Turing Institute}
  \city{}
  \country{}
}
\email{sa894@cam.ac.uk}

\author{Eiko Yoneki}
\affiliation{%
  \institution{University of Cambridge, UK}
  \city{}
  \country{}
}
\email{eiko.yoneki@cl.cam.ac.uk}

\begin{abstract}
  Current auto-tuners struggle with computer systems due to their large complex parameter space and high evaluation cost.
We propose  \BoGraph{}, an auto-tuning framework that builds a graph of the system components before optimizing it using causal structure learning.
The graph contextualizes the system via decomposition of the parameter space for faster convergence and handling of many parameters.
Furthermore, \BoGraph{} exposes an API to encode experts' knowledge of the system via performance models and a known dependency structure of the components.
We evaluated \BoGraph{} via a hardware design case study achieving $5x-7x$ improvement in energy and latency over the default in a variety of tasks.

\end{abstract}

\maketitle

\tolerance=1000 \hbadness=1000

\section{Introduction}
Computer systems facilitate the execution of a wide range of diverse workloads by exposing tunable configurations to meet users' demands.
Examples of these configurations are hardware designs \cite{Gem5Simulatorbinkert2011} and database knobs \cite{InquiryMachineLearningbasedvanaken2021}.
Tuning systems configuration is tedious due to the many parameters involved and the long evaluation time.
For example, in our case study on hardware design optimization, there are $2^{64}$ unique designs. % (see \autoref{tab:params}).
This complexity necessitates for auto-tuners to aid practitioners.
However, current research into auto-tuners focuses on optimizing neural network hyperparameters \cite{MicrosoftNniopensource2020, RandomSearchHyperparameterbergstra2012, PracticalBayesianOptimizationsnoek2012} and encounters significant challenges when applied to tuning computer systems specifically.

\subsection{\BoGraph{}}
\BoGraph{} builds on Structured Bayesian Optimization (SBO) \cite{Dalibard2017} principles to contextualize the system and finds optimal system configurations in a few trials.
SBO requires experts to encode their system knowledge using hand-crafted probabilistic models.
\BoGraph{} differentiates itself from previous works by simplifying the process of using SBO.
It discovers probabilistic models from logs using its novel structure discovery pipeline.
\BoGraph{}'s API allows experts to encode contextual knowledge of the system, such as models or dependencies between components, for a more robust and interpretable auto-tuner.
Next, we summarize the challenges \BoGraph{} targets in optimizing computer systems.

\subsection{Challenges in systems optimization}\label{sc:challenges}
\subsubsection{Evaluation is slow and expensive} \label{challenge:expensive}
Evaluating a single configuration on a system takes several minutes (see \autoref{fig:env_time}) to several hours \cite{Floratou2017, InquiryMachineLearningbasedvanaken2021, rlDevicePlacement}.
This is costly especially when optimizing cloud instances \cite{Alipourfard2017} or databases \cite{InquiryMachineLearningbasedvanaken2021}.
Hence, the need to find optimized configurations with the fewest evaluations.
As a result, tuners that rely on many iterations are too costly to use, e.g., reinforcement learning \cite{ReinforcementLearningIntroductionsutton1998}, random search \cite{RandomSearchHyperparameterbergstra2012},
evolutionary search \cite{Ansel2010}, hill-climbing \cite{Ansel2014}, or population-search \cite{PopulationBasedTrainingjaderberg2017}.
\BoGraph{} leverages all available information from the system logs and captures it using Gaussian Process (GP) \cite{GaussianProcessesMachinerasmussen2008} a sample-efficient model to contextualize the system.
Additionally, it allows incorporating existing experts' knowledge to speed up the process further.

\subsubsection{Large number of parameters} \label{challenge:large_param}
\begin{figure}
    \centering
    \begin{subfigure}[b]{.225\textwidth}
        \includegraphics[width=\textwidth,keepaspectratio]{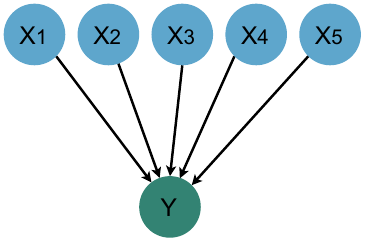}
        \caption{The model has to consider all the parameters jointly.}
        \label{fig:decomposed_a}
    \end{subfigure}
    \hfill
    \begin{subfigure}[b]{.225\textwidth}
        \includegraphics[width=\textwidth,keepaspectratio]{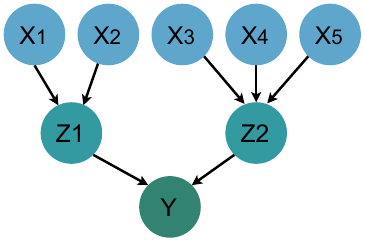}
        \caption{Decomposition reduces the effective dimensions naturally.}
        \label{fig:decomposed_b}
    \end{subfigure}
    % \vspace{3mm}
    \caption{The structure reduces the effective dimensions.}
    \label{fig:decomposition}
    \Description{
        The Figure shows the system decomposability.
        On the left is a standard optimization approach that considers all parameters jointly, with a max of five dimensions.
        On the right, a decomposed method breaks down optimization along the available metrics line, leading to fewer dimensions considered for each optimization target.
    }
\end{figure}

Using GPs comes at the cost of increased computational complexity and difficulty to scale to $D > 10$ parameters without sacrificing their efficiency  \cite{WhenGaussianProcessliu2019}.
Standard Bayesian Optimization \cite{PracticalBayesianOptimizationsnoek2012} methods often rely on GPs, which do not scale to many parameters.
\BoGraph{} enables the GPs to scale by using the system's dependency graph to naturally reduce each model dimension as illustrated in \autoref{fig:decomposition}.
\subsubsection{Complex parameters dependency} \label{challenge:dependency}
Computer systems parameters often have a complex relationship; a parameter can depend on another parameter, e.g. the memory's structure and size.
This relationship is non-linear and difficult to learn using traditional methods.
\BoGraph{} exposes an API for experts to encode a parametric probabilistic model that captures parameters' complex relationship.

\subsubsection{Designing bespoke models is complicated} \label{challenge:perf_model}
Hand-crafting models that capture the system's complexity requires system and machine learning expertise.
Furthermore, validating and evaluating the models is a trial-and-error process.
Making tuners that rely on expert models such as BOAT \cite{Dalibard2017} and Causal Bayesian Optimization \cite{CausalBayesianOptimizationaglietti2020} difficult to use.
\BoGraph{} allows encoding models if they exist, and it provides a pipeline that learns these models if necessary.
It captures statistical dependency between parameters and system metrics then uses GPs to model their interactions.

\subsection{Contributions}
We utilized  \BoGraph{} to optimize the design space ($\approx 2^{64}$) of an accelerator simulator, gem5-Aladdin \cite{CodesigningAcceleratorsSocshao2016, Gem5AladdinSoCSimulatorharvard-acc2016}.
\BoGraph{} learned the system's components dependencies from logs, then modeled them using a graph of probabilistic models.
\BoGraph{} found configurations in twenty iterations that took the next best auto-tuner a hundred iterations.
Moreover, while improving the energy-latency utilization by $5-7x$ over the default, no other method came close to its performance.

We summarize our contributions as follows:
\begin{itemize}
    \item An auto-tuner for computer systems that handle many parameters and find optimal configurations quickly.
    \item A SBO framework automatically learns a graph of probabilistic models from the system's logs.
    \item An API for experts to express the system's dependencies or encode probabilistic models of its components.
    \item Optimized all the accelerator design choices in gem5-Aladdin and improved energy-latency by $5-7$x.
\end{itemize}

\section{Background}
\subsection{Bayesian optimization}
\textit{Bayesian Optimization} (BO) \cite{TakingHumanOutshahriari2015} is an iterative sample-efficient method to optimize configurations.
Formulated as $x^* = argmin(\alpha(\mathbb{M}(f))$, where $x^*$ is the optimal configuration,
$\alpha$ is the acquisition function, $\mathbb{M}$ is a probabilistic model of the system, and $f$ is the system.
Its efficiency comes from generating samples using $\mathbb{M}$ that $\alpha$ optimizes to find configurations that improve the system without direct interaction with it.
The choice of $\mathbb{M}$ and  $\alpha$ directly impact the required evaluations to find optimal configurations.

\subsection{Gaussian Process}
Gaussian Process (GP) \cite{GaussianProcessesMachinerasmussen2008} is a flexible sample-efficient model commonly used in BO.
GP captures the impact of the configurations against the objective in the mean function and co-variance in its kernel.
In practice, GP struggles with problems that have $D > 10$ \cite{BayesianOptimizationBillionwang2016} as it scales cubically to the data and dimensions.
Its runtime requires an expensive matrix inversion operation for training and sampling, hence its scalability issues.
The computational complexity of the GP renders it ineffective for high-dimensional problems.

\subsection{Probabilistic DAG}
\BoGraph{} models the system using a directed-acyclic-graph (DAG), where the nodes are probabilistic models that capture a component of the system, and the edges are the conditional dependency between the models.
This representation, a DAG of probabilistic models, is known as Bayesian Network \cite{ProbabilisticGraphicalModelskoller2009}.
Bayesian Network has the advantage of factorizing the joint distribution of a node into a local distribution that depends only on the parent:
$P(X; G) = \prod_{j=i}^{p}P(X_i|X_{parent(i)})$ where $X_i$ is the \textit{ith} node in the graph and \textit{parent(i)} is the direct parent of the node.
Nodes factorization circumvents \textit{the curse of dimensionality} and allows each node to scale independently.

\section{ \BoGraph{} Framework}
\begin{figure}
  \centering
  \includegraphics[width=1\linewidth,keepaspectratio]{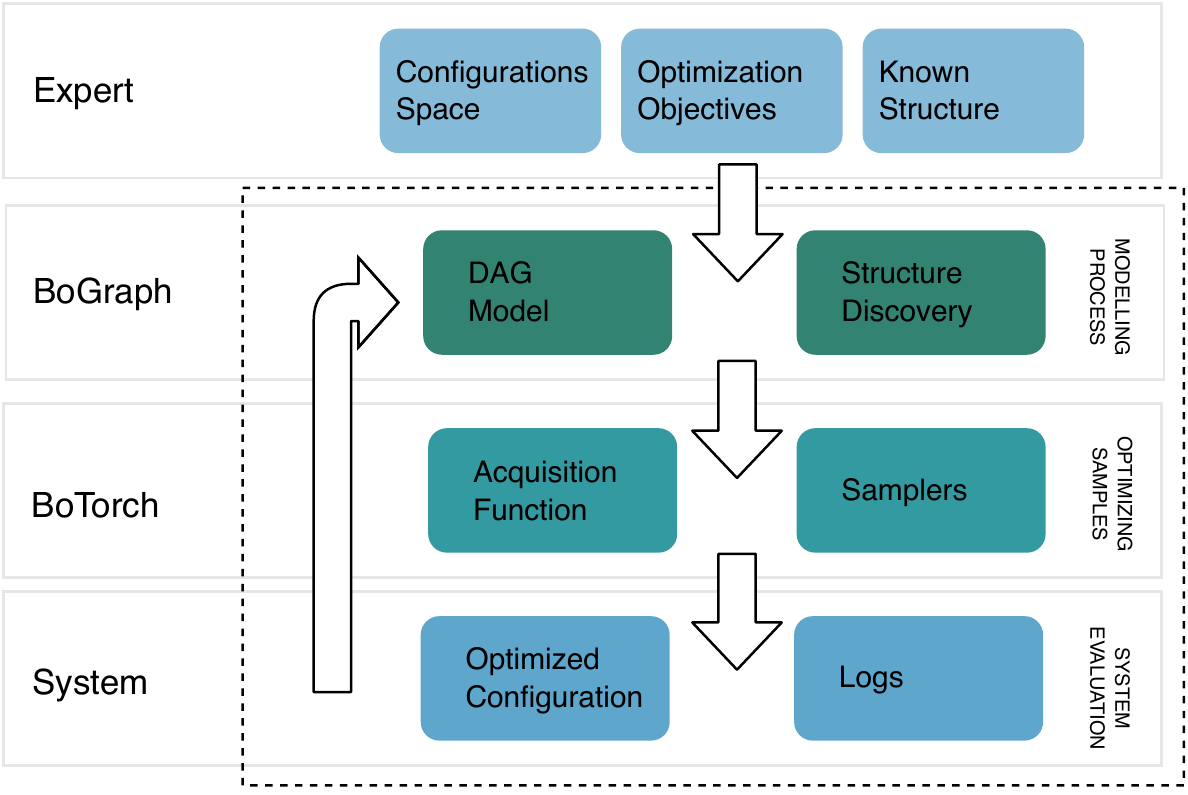}
  \caption{
    \small
    \BoGraph{}'s optimization loop. The expert provides specifications of the problem. Then,  \BoGraph{} learns a system structure from the logs and combines expert knowledge to build a probabilistic model. Finally, it generates samples from the model and optimizes them to propose optimized configurations to the system.
    \label{fig:bo_loop}
  }
\end{figure}
\BoGraph{} is an auto-tuning framework for systems with many configurations.
It builds on Structured Bayesian Optimization (SBO) \cite{Dalibard2017} by utilizing contextual knowledge to converge quickly.
\BoGraph{} novelty is to allow a hybrid method for injecting expert knowledge into the system as well as learning system structure from logs.
Using its pipeline, it identifies statistical links between the system components and uses them to build a system's probabilistic DAG.

\autoref{fig:bo_loop} shows  \BoGraph{}'s main components.
The expert encodes any knowledge of the system and the target objectives.
Then,  \BoGraph{} scans the system logs to learn the statistical dependency of the system's components.
Next, it combines the expert's knowledge with the learned dependency producing a probabilistic DAG.
Finally, it suggests configurations to evaluate by generating many samples from the model and optimizing them using an acquisition function.
Its modular design facilitates the use of any sampler and acquisition functions, with BoTorch \cite{BoTorchFrameworkEfficientbalandat2020} being the default.

\subsection{Structure discovery from logs} \label{sc:struct_learning}
\begin{figure}
  \centering
  \includegraphics[width=0.75\linewidth, keepaspectratio]{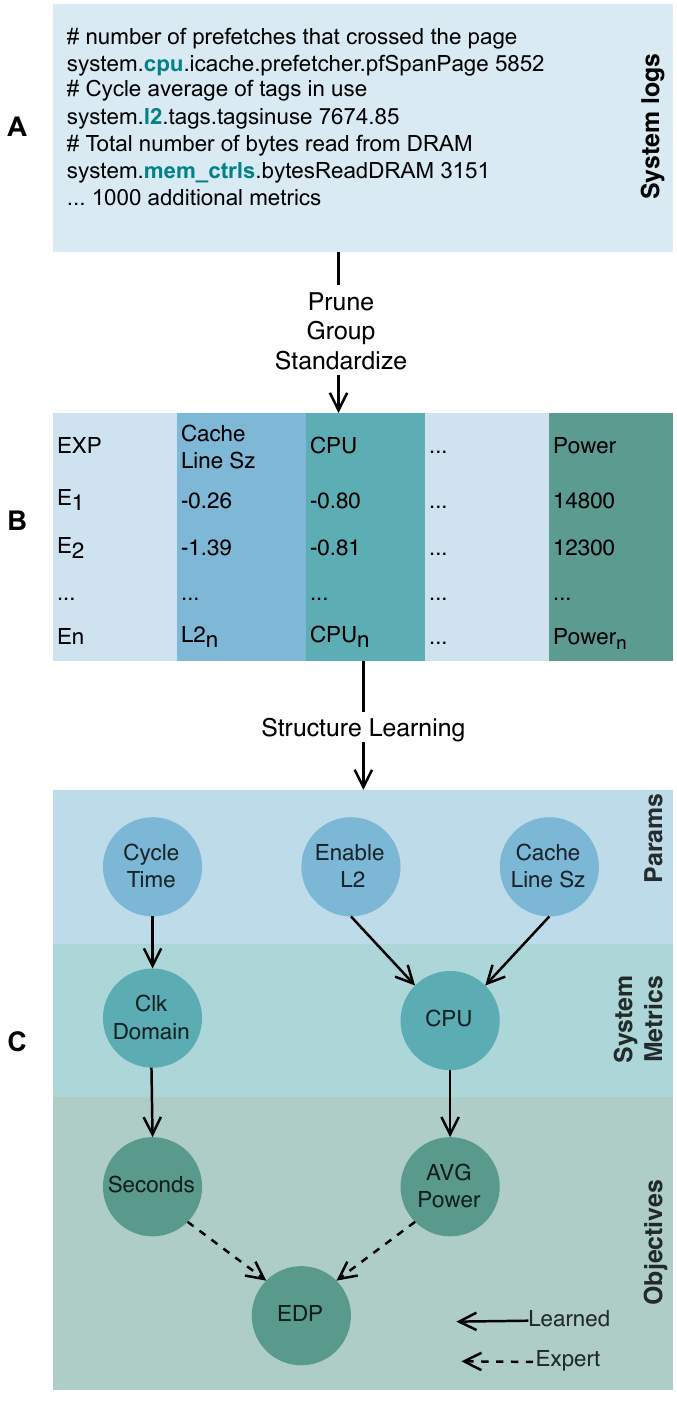}
  \caption{
    \small
    A) A snippet of gem5's (the environment) logs.
    B) The logs after summarizing and grouping.
    C) The structure resulted by calculating a statistical correlation between components (respecting parameters-component causality) and applying the expert-defined edges (the entire graph is too large for the paper).
    \label{fig:bograph_all}
  }
\end{figure}

\BoGraph{}'s first component is the structure discovery process shown in \autoref{fig:bograph_all} that learns a dependency DAG from the system logs.
The pipeline first summarizes the logs, then discovers the statistical dependency using structure learning \cite{ProbabilisticGraphicalModelskoller2009, DagsNoTearszheng2018},
and finally connects any expert-defined edges to produce a dependency DAG of the system components.

% \vspace{-2mm}
\subsubsection*{Summarizing the logs}
Computer systems generate logs used for monitoring and debugging by experts \cite{SiteReliabilityEngineeringbeyer2016}, \BoGraph{} uses the logs to learn components dependencies.
However, the logs need to be summarized first as there are often many log entries causing \textit{the curse of dimensionality}.
For example, in the case study described in \autoref{sc:soc_eval}, gem5-Aladdin reports more than $1000$ log entry (\autoref{fig:bograph_all}A), \BoGraph{} pipeline summarizes them to $7$ entries (\autoref{fig:bograph_all}B).

\subsubsection*{Preprocessing}
First, \BoGraph{} removes uninformative metrics by pruning ones with low variance between executions.
Then, they are standardized to ensure that different measuring units do not skew the learning algorithm.
Then, \BoGraph{} exploits the structured property of logs to summarize them.
The logging protocol encodes where the log was generated in the system \cite{CharacterizingLoggingPracticesyuan2012, WhereDevelopersLogfu2014}.
For example, in the case study on gem5, the structure is \textit{\textcolor[HTML]{02818A}{Component}.SubComponent.Measurement}, as shown in \autoref{fig:bograph_all}A, \BoGraph{} groups all measurements under \textit{SubComponent} in one group as they influence the same component in the system, then they are compressed into a single statistic using Factor Analysis (FA) \cite{EasyGuideFactorkline2014}.
This compression works because logs generated at the \textit{SubComponent} level are dependent on each other.
The user can decide on the granularity of the grouping based on their need.
A fine granularity is helpful to understand the system, but it is more expensive to evaluate.
We used the highest level of granularity which mapped gem5 to six main categories: L2, Datapath, Mem Ctrls, Membus, Tol2Bus, and CPU.
This process applies to all system logs, including previous system evaluations.
At the end of this stage,  \BoGraph{} produces a summary of the logs as seen in \autoref{fig:bograph_all}B for the structure learning algorithm.

\subsection{Structure learning}
Defining systems' structure is an error-prone process; hence the need to learn it from data, an active research area in the Bayesian community \cite{SurveyBayesianNetworkscanagatta2019}.
These methods propose several DAGs and score them based on a trade-off between data fit and the complexity of the graph.
\BoGraph{} uses a structure learning algorithm, NoTears \cite{DagsNoTearszheng2018}. NoTears approximates the DAG score using a differentiable function, then minimizes its error, which is easily parallelizable on GPUs \cite{Causalnex2021}.
It is simple to extend \BoGraph{} with other structure learning algorithms by implementing the \BoGraph{}'s StructureLearning interface.
We encode a set of restrictions in these algorithms that ensures the system's parameters are not allowed to have parents, and the objectives nodes are not allowed to have children to ensure the causal flow of the optimization task.
Then apply any expert-defined edges resulting in \autoref{fig:bograph_all}C.
Optionally, \BoGraph{} takes an update scheduler that decides when to perform structure learning.
By default, it performs structure learning every quarter of the optimization budget.

\subsection{Graph of a Probabilistic Models}
\begin{figure}
  \centering
  \includegraphics[width=1.\linewidth, keepaspectratio]{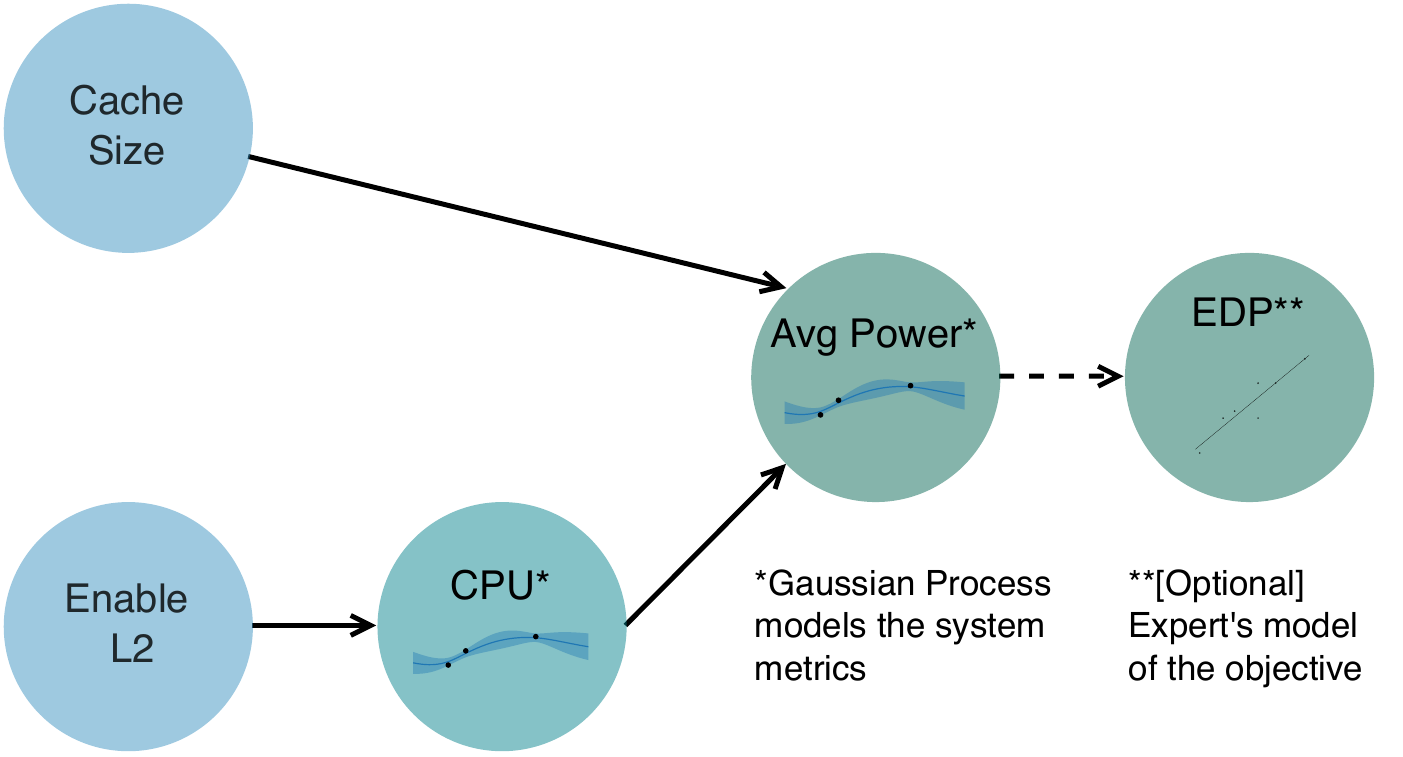}
  \caption{
    \small
    \BoGraph{} maps the node in the structure to probabilistic models such as GPs or expert's models, creating probabilistic DAG.
    \label{fig:structure}
  }
\end{figure}
\BoGraph{} builds the probabilistic DAG by placing a GP node on intermediate nodes, and then it connects any expert-defined models as illustrated in \autoref{fig:structure}.
The probabilistic DAG nodes' are \textit{compartmentalized}, where each node predicts a single value conditioned on its parents directly and can be trained independently.
As a result, any node can be an optimization target, providing free multi-objective modeling and enabling a parallelizable training.
We will investigate both in a follow-up project.

\begin{algorithm}[tb]
  \small
  \caption{Sampling the probabilistic DAG}
  \label{alg:dag_sampling}
  \begin{algorithmic}
    \STATE {\bfseries Input:} sampler $s$, DAG model $M$, system $SY$
    \STATE initialize a samples cache: $Q := Dict()$
    \STATE get nodes along the path to the objectives: $P = M.path(SY.obj)$
    \FOR{node $n$ {\bfseries in} $P.nodes$}
    \IF{$n \in SY.params$}
    \STATE sample parameters: $p_{samples} = s.sample(n.space())$
    \STATE cache generated samples: $Q \cup (n.key,p_{samples})$
    \ELSE
    \STATE condition on parents: $C = n.predict(Q[n.parents])$
    \STATE cache node results conditioned on parents: $Q \cup (n.key, C)$
    \ENDIF
    \ENDFOR
    \STATE {\bfseries Return:} $Q[SY.objectives]$
  \end{algorithmic}
\end{algorithm}

\textbf{Sampling process.}
Allowing general probabilistic models comes at the cost that closed-form acquisition function optimization is not possible; instead,  \BoGraph{} samples the model and then optimizes over the samples.
This design gives it the freedom to use any probabilistic model inside the graph, such as Bayesian Neural Networks \cite{BayesianNeuralNetworksmackay1995}.
Algorithm \ref{alg:dag_sampling} shows the process to generate a prediction and samples from a probabilistic DAG.
First,  \BoGraph{} finds all the nodes with a direct path from the parameters to the objective.
Then, it creates a cache to hold all the samples and avoid recomputing them for nodes with multiple children.
Next, each node condition on the parents' prediction of the samples, and finally output the posterior of the objectives.

\subsection{Structured Bayesian Optimization}
\begin{algorithm}
  \small
  \caption{
    \label{alg:bo_loop}
    \BoGraph{} Bayesian Optimization loop}
  \begin{algorithmic}
    \STATE {\bfseries Input:} acquisition function $\alpha$, system $E$, budget
    \REPEAT
    \STATE build DAG model from logs: $DAG_M =  \BoGraph{}{}(E.logs())$
    \STATE select next to point to evaluate: $x_{n+1} = \alpha(s.sample(DAG_M))$
    \STATE submit $x_{n+1}$ to be evaluated: $E.update(x_{n+1})$
    \UNTIL{$budget = 0$}
  \end{algorithmic}
\end{algorithm}
% \vspace{-2mm}

\textbf{Optimization loop.}
Algorithm \ref{alg:bo_loop} details the optimization loop.
\BoGraph{}'s probabilistic DAG combines several models' posterior, producing a complicated posterior that cannot be optimized directly.
Instead, we use quasi-acquisition functions that optimize the estimate from samples rather than the model directly.
BoTorch \cite{BoTorchFrameworkEfficientbalandat2020} provides a suite of utility functions for BO frameworks, including quasi-acquisition functions and samplers.
We use BoTorch's Sobol sequence sampler to generate diverse samples of the posterior, then use the quasi-Expected Improvement acquisition function to optimize the estimates and produce optimization candidates.

\subsection{ \BoGraph{} API}
\BoGraph{} API streamlines the process of expressing system designers' knowledge.
Listing \ref{lst:API} shows several of the functions  \BoGraph{} that enable experts to encode their knowledge in two ways:
The first is a DAG API based on \textit{networkX} \cite{ExploringNetworkStructurehagberg2008} to express the dependencies between parameters and metrics supporting many methods to store and load DAGs.
The second is an advanced API that expresses experts' knowledge using \textit{Pyro} \cite{PyroDeepUniversalbingham2018}, a probabilistic programming language.
By default,  \BoGraph{} uses a GP from GPyTorch \cite{GPyTorchBlackboxMatrixMatrixgardner2019} a performant GP implementation as the probabilistic model if no model is defined.
The user can override the kernel of the GP to inject any prior information about the objective function distribution; this is only recommended for advanced users.

% frame=lines,
% linenos,
\begin{listing}
  \begin{minted}
        [
        fontsize=\scriptsize,
        framesep=1mm,
        escapeinside=||,
        autogobble=true,
        frame=lines,
        ]
    {python}

class ExpertModel(|\textcolor{red}{PyroNode}|): # User extends PyroNode to build models
  def model(x1: Tensor, obs: Tensor = None):
    i = pyro.sample("intercept", dist.Uniform(0., 5.))
    linear_mean = i + pyro.sample("bias_x1", dist.Normal(0.,1.))
    ... # a standard Bayesian model using Pyro
prior_dag = | \BoGraph{}{}|Prior() # | \BoGraph{}{}| uses `networkx` API 
prior_dag.add_edges_from(['x1', 'f1(x1)']) # f1(x1) depends on x1
prior_dag.add_model("f1(x1)", ExpertModel()) # add expert's model 
prior_dag.add_model("y", GPyTorchModel()) # standard GP model
prior_dag.add_tabu_edge("x1", "y") # Edges known to not be correlated
|\BoGraph{}{}|.optimize(objective=Forrester2D, prior=prior_dag)
    \end{minted}
  \vspace{-2mm}
  \caption{ \BoGraph{} API showing expert knowledge injection}
  \label{lst:API}
\end{listing}
\subsection{Summary}
\BoGraph{} simplifies the process of integrating SBO in systems for faster optimizer convergence.
The main loop performs structure learning of the system's components by processing the logs and transforming it into a probabilistic model.
\BoGraph{} exposes an API that allows experts to encode dependency structure and probabilistic models.

\section{Evaluation}\label{sc:soc_eval}
System-on-a-chip (SoC) designers often experiment on a simulator \cite{Gem5Simulatorbinkert2011, AladdinPrertlPowerperformanceshao2014, EyerissV2Flexiblechen2019} to reduce the energy and latency of integrated chips \cite{ExploringExplorationTutorialpimentel2016, DeterminingOptimalCoherencybhardwaj2019}.
We optimized  Energy-Delay Product (EDP)\cite{EnergyperformanceTradeoffsProcessorazizi2010} of gem5-Aladdin \cite{CodesigningAcceleratorsSocshao2016}
where $EDP = energy * (\frac{1}{sim\_seconds})^2$ by tuning all the design parameters in \autoref{tab:params}.
\begin{table}
    \caption{
        \label{tab:params}
        \small
        The gem5-aladdin parameters from their sweeper \cite{Gem5AladdinSoCSimulatorharvard-acc2021}.
        The $ \approx 2^{64}$ unique design is enormous hence the need for \BoGraph{}.
        The parameters spaces in red are pruned using a power function.
    }
    % \vskip 0.15in
    \vspace{-2mm}
    \begin{center}
        \begin{small}
            \begin{sc}
                \begin{tabular}{l r r}
                    \toprule
                    Parameter                                    & Min  & Max   \\
                    \midrule
                    Cycle Time                                   & 1    & 10    \\
                    Pipelining                                   & 0    & 1     \\
                    Cache Size \textcolor{red}{(power of two)}   & 14   & 17    \\
                    Cache Assoc \textcolor{red}{(power of two)}  & 0    & 5     \\
                    Cache Line Sz\textcolor{red}{(power of two)} & 5    & 7     \\
                    Cache Hit Latency                            & 1    & 5     \\
                    Cache Queue Size                             & 16   & 256   \\
                    Cache Bandwidth                              & 2    & 18    \\
                    Tlb hit latency                              & 1    & 32    \\
                    Tlb miss latency                             & 1    & 32    \\
                    Tlb entries \textcolor{red}{(power of two)}  & 0    & 5     \\
                    Tlb assoc \textcolor{red}{(power of two)}    & 0    & 5     \\
                    Tlb page size                                & 1024 & 32768 \\
                    Tlb max outstanding walks                    & 2    & 16    \\
                    Tlb bandwidth                                & 2    & 16    \\
                    L2cache Size \textcolor{red}{(power of two)} & 17   & 21    \\
                    Enable L2                                    & 0    & 1     \\
                    Pipelined Dma                                & 0    & 1     \\
                    Ready Mode                                   & 0    & 1     \\
                    Ignore Cache Flush                           & 0    & 1     \\
                    \bottomrule
                \end{tabular}
            \end{sc}
        \end{small}
    \end{center}
    \vskip -0.1in
\end{table}

\subsubsection*{Evaluation goals} We evaluated the following claims:
\begin{itemize}
  \item Executing \BoGraph{} is faster than a full system evaluation, \ref{sc:exe_time}.
  \item \BoGraph{} converges towards optimal configurations faster than other state-of-the-art methods, \ref{sc:edp_improv}.
  \item \BoGraph{} produces interpretable system structures that reduces the maximum dimension, \ref{sc:structure_discovery}.
\end{itemize}

\subsubsection*{Setup}
The experiment ran on RTX3060TI, Intel i7 8700k, and 32GB RAM.
BoGraph used GPs from GPyTorch v1.2 \cite{GPyTorchBlackboxMatrixMatrixgardner2019}, samplers and acquistion functions from BoTorch v0.5 \cite{BoTorchFrameworkEfficientbalandat2020}, and CausalNex v0.11 \cite{Causalnex2021} for structure learning.
The gem5-Aladdin simulator was compiled using MESI Two Level Aladdin protocol with MathExprPowerModel as power module.
\subsubsection*{Benchmark}
Used a mix of data and compute workloads from MachSuite \cite{MachsuiteBenchmarksAcceleratorreagen2014} as is common in SoC literature \cite{CodesigningAcceleratorsSocshao2016, ExploringExplorationTutorialpimentel2016}.
\subsubsection*{Expert knowledge.} The EDP formula.

\subsubsection*{Baselines}
We used these baselines in the experiments:
\begin{itemize}
  \item \textbf{Default} \cite{Gem5AladdinSoCSimulatorharvard-acc2016}: the settings provided by the simulator.
  \item \textbf{Random} \cite{RandomSearchHyperparameterbergstra2012, MicrosoftNniopensource2020}: useful when small subset of the dimensions significantly impact the objective.
  \item \textbf{PBTTuner} \cite{PopulationBasedTrainingjaderberg2017a, MicrosoftNniopensource2020}:
        a bag of model approach to finding a collection of optimal configurations.
  \item \textbf{BoTorch} \cite{BoTorchFrameworkEfficientbalandat2020}:
        shows the effectiveness of \BoGraph{} as they use the same samplers and acquisition functions.
  \item \textbf{DeepGP}  \cite{DeepGaussianProcessesdamianou2013}: Provides unsupervised structure discovery in the latent space.
        We configured a two-layer GPs with 64 inducing points (Our GPU's max VRAM).
        % \item \textbf{DeepGP}  \cite{DeepGaussianProcessesdamianou2013}:  two-layer GPs with 64 inducing points, unsupervised structure discovery in the latent space.
\end{itemize}

\subsection{Execution time} \label{sc:exe_time}
\begin{figure}[b]
  \begin{subfigure}[t]{\linewidth}
    \centering
    \includegraphics[height=120pt, keepaspectratio]{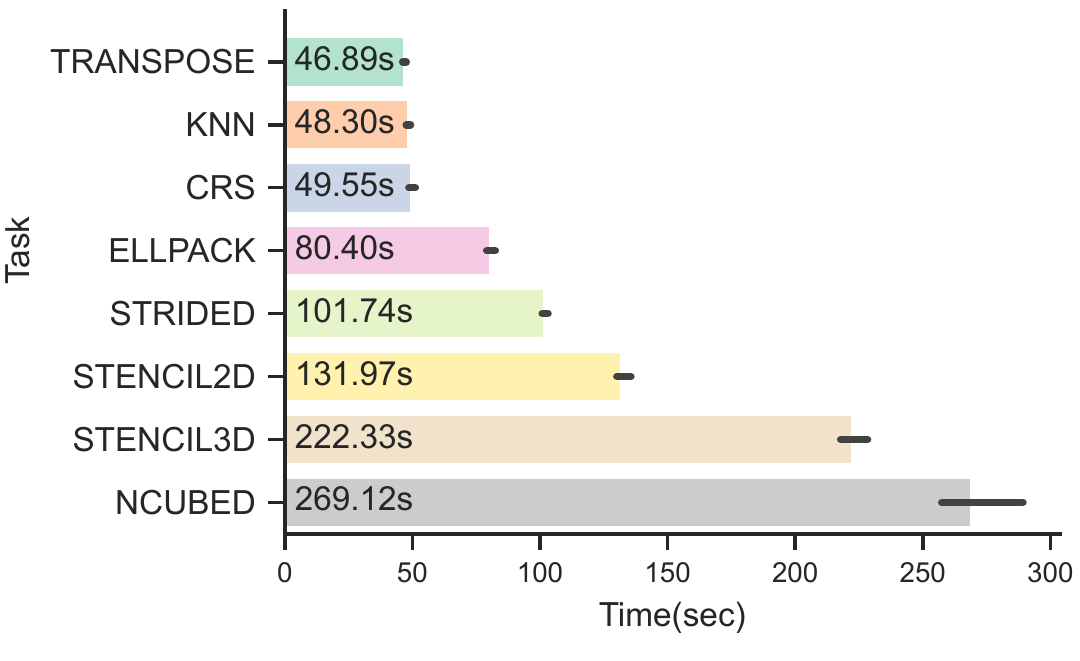}
    \caption{
      \small
      Tasks' average execution time, with variance from $1800$ runs.
    }
    \label{fig:env_time}
  \end{subfigure}
  \hfill
  \begin{subfigure}[t]{\linewidth}
    \centering
    \includegraphics[height=140pt, keepaspectratio]{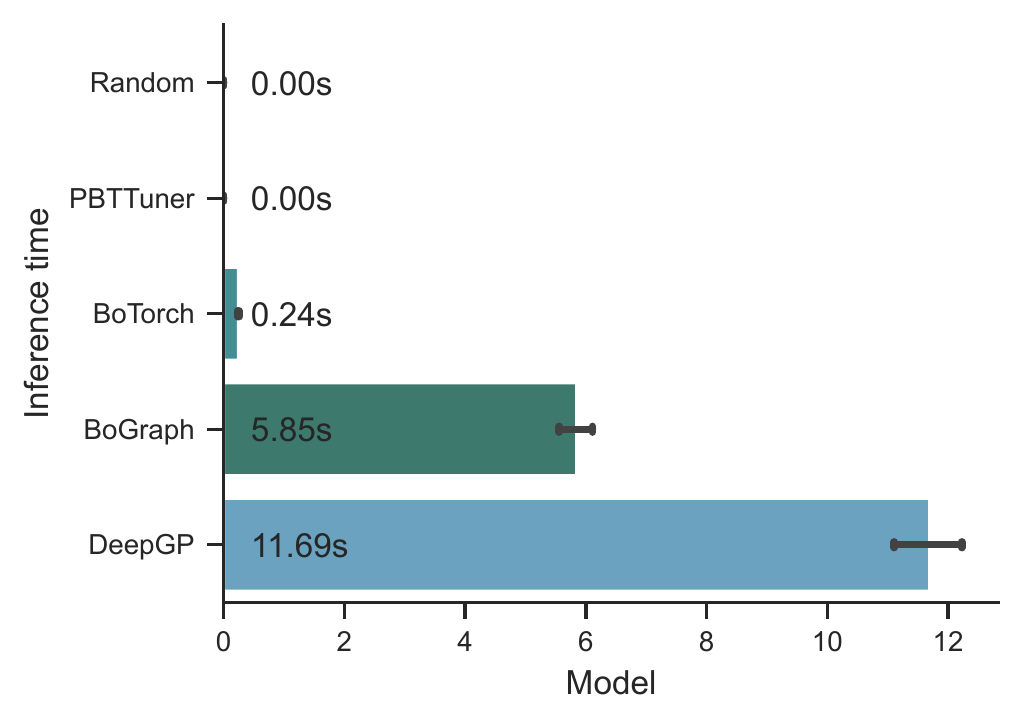}
    \caption{
      \small
      Auto-tuners' average time to train and propose a configuration, with variances from $300$ runs.
    }
    \label{fig:model_time}
  \end{subfigure}
  \caption{
    \small
    Optimizers' execution time is dominated by the gem5-Aladdin task execution time.
    \label{fig:env_vs_model_time}
  }
\end{figure}

We compared the execution time of the simulator for each task and compared it to the overhead of using an auto-tuner in \autoref{fig:env_vs_model_time}.
The results show that the optimizers are cheaper than a full system evaluation.
Hence the need for auto-tuners that perform better than simple parameter sweeps.

% \vspace{-2mm}
\subsection{EDP Optimization}\label{sc:edp_improv}
\begin{figure*}
  \centering
  \includegraphics[width=0.65\linewidth, keepaspectratio]{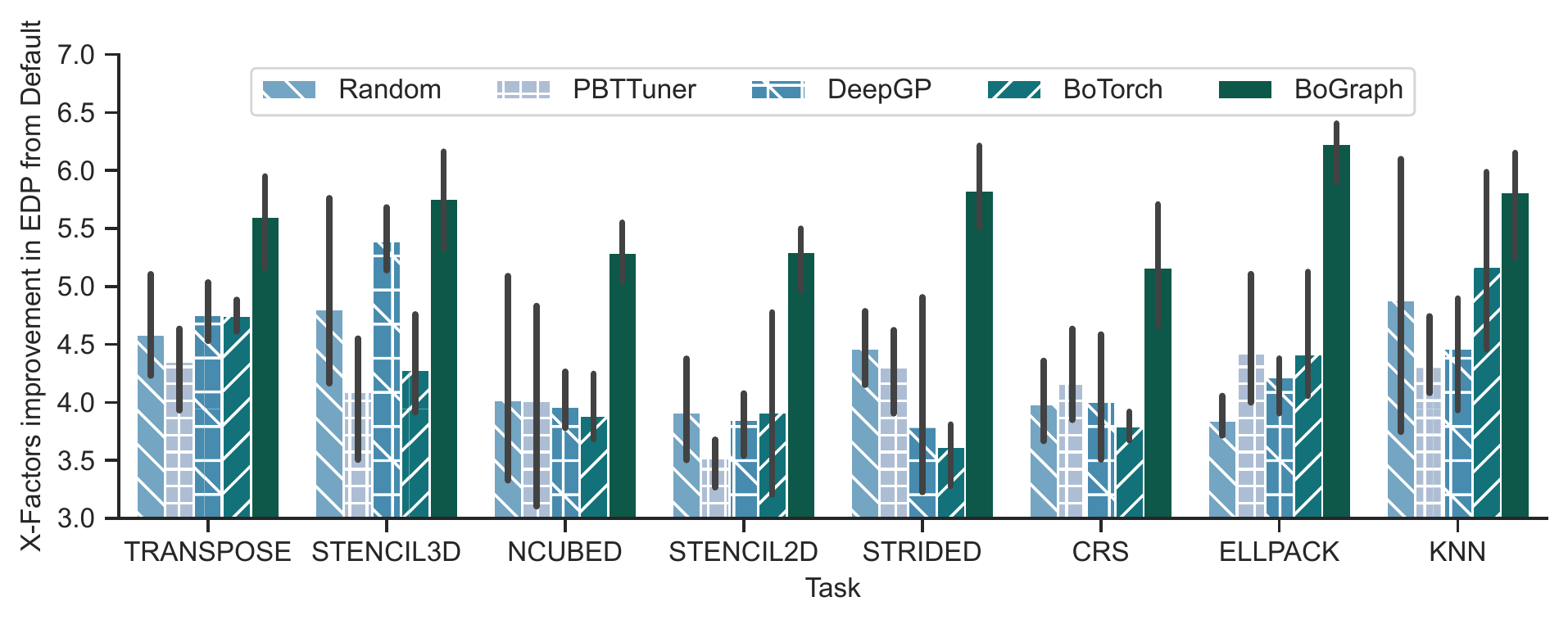}
  \vspace{-4mm}
  \caption{
    \small
    The best found EDP in $100$ steps.
    The y-axis shows X-factors improvement over default settings and the variance of three runs.
  }
  \label{fig:all_per}
\end{figure*}

\autoref{fig:all_per} examines the improvement in the EDP from the default configurations across all tasks.
\BoGraph{} outperformed every other method and improved over the default by $5x-7x$ factors.
Examining that further, \autoref{tab:ellpack_perf} shows that  \BoGraph{} found configurations with the least power consumption for the ELLPACK task.
Other tuners are stuck in local minima that do not improve the energy as they are unaware the EDP is impacted by power.
\BoGraph{} is aware that the latency and power impact EDP thus can avoid the local minima.
The same story applies to the other tasks in the benchmark.

\begin{table}
  \caption{
    \label{tab:ellpack_perf}
    \small
    ELLPACK's performance for the median best configuration proposed by the models, lower values are better.
  }
  \begin{center}
    \begin{small}
      \begin{sc}
        \begin{tabular}{l|rrr}
          \toprule
          model               & log EDP                    & Latency                   & Power (mWatts)             \\
          \midrule

          \textbf{\BoGraph{}} & ${\mathbf{8.12 \pm 0.28}}$ & $\mathbf{11.06 \pm 0.19}$ & $\mathbf{27.37 \pm 11.59}$ \\
          DeepGP              & $10.14 \pm 0.27$           & $11.06 \pm 3.92$          & $203.15 \pm 46.34$         \\
          PBTTuner            & $10.34 \pm 0.60$           & $10.81 \pm 4.46$          & $264.36 \pm 147.79$        \\
          BoTorch             & $10.44 \pm 0.62$           & $11.06 \pm 3.90$          & $108.46 \pm 102.97$        \\
          Random              & $10.74 \pm 0.19$           & $18.32 \pm 4.81$          & $143.94 \pm 194.83$        \\
          \midrule
          Default             & $14.50 \pm 0$              & $94.80 \pm 0$             & $220.34 \pm 0$             \\
          \bottomrule
        \end{tabular}
      \end{sc}
    \end{small}
  \end{center}
\end{table}

\subsubsection*{Convergence}
\begin{figure}
  \centering
  \includegraphics[width=0.8\linewidth, keepaspectratio]{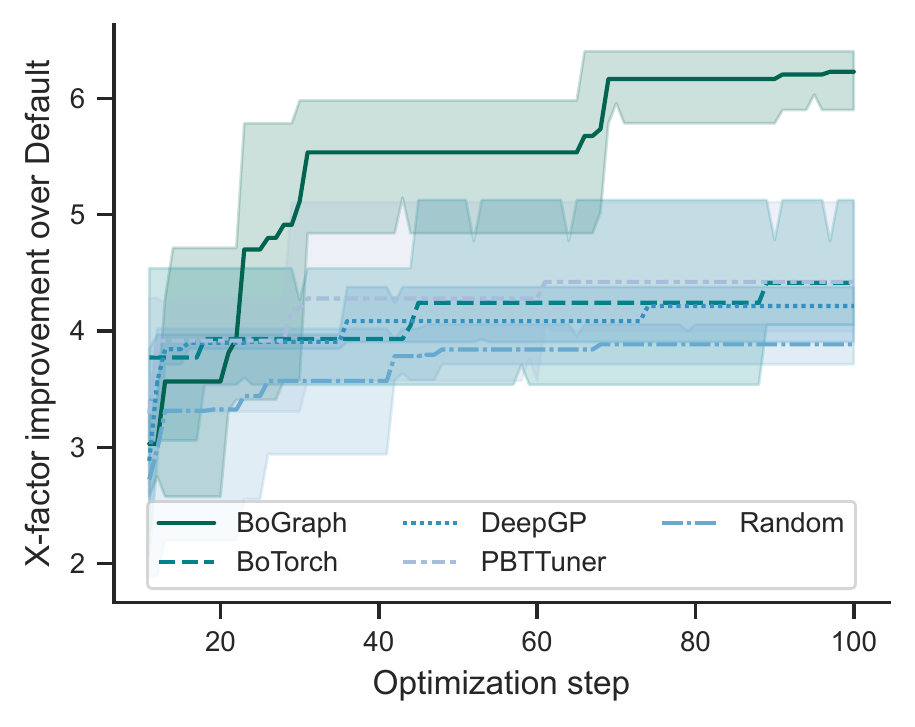}
  \vspace{-2mm}
  \caption{
    \small
    Auto-tuners' improvement over default, with a median best configuration and the error bars showing a minimum of 3 runs.
    \label{fig:edp_convergence}
  }
\end{figure}
\autoref{fig:edp_convergence} examines the convergence rate of each optimizer as minimizing the evaluations reduces the cost significantly (Challenge \ref{challenge:expensive}).
\BoGraph{} reached the best configuration the quickest and continued to improve while other methods plateaued.
Confirming that tuning computer systems is challenging for most auto-tuners.
\BoGraph{} improvement jumps at every 20th step in \autoref{fig:edp_convergence} are caused by the structure learning algorithm, as its update frequency defaults to a quarter of the evaluation budget.

\subsection{Structure discovery}\label{sc:structure_discovery}
\begin{figure}
  \centering
  \includegraphics[width=0.8\linewidth, keepaspectratio]{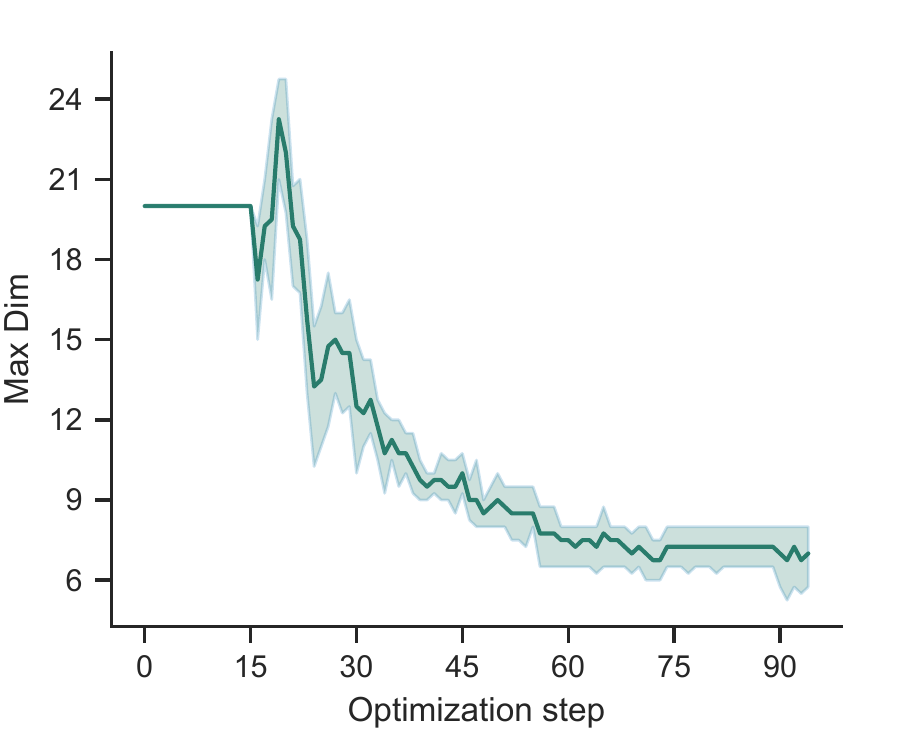}
  \vspace{-2mm}
  \caption{
    \small
    \BoGraph{} DAG max-dimension (lower is better) of three repeated runs.
    The max dimension is defined as $MD = \forall_{n \in G} max(indegree(n))$, each node becomes a probabilistic model.
    \label{fig:structure_reduce_dim}
  }
\end{figure}

There are over $1000$ metrics reported in a gem5 statistics file.
\BoGraph{} shields the end-user from the difficult task of defining a probabilistic model while providing an insightful structure to the designer as shown in \autoref{fig:bograph_all}C.
Furthermore, \autoref{fig:structure_reduce_dim} shows that learning structure does reduce the maximum dimension for the models in the graph.

\section{Oppurtinities and limitation.}
\subsection{Oppurtinities}
\BoGraph{} is still evolving, with aims to apply to different problem spaces, evaluate it on a multi-objective problem, and improve its computational engine.
\subsubsection*{Multi-objective optimization.}
Multi-objective modeling is an interesting opportunity for computer system objectives often compete (latency vs. throughput, energy vs. latency).
Modeling multiple objectives is trivial in \BoGraph{} as a result of using DAGs as an internal data structure as any node can be sampled.
Then using off-the-shelf multi-objective utility functions such as expected hypervolume improvement function \cite{DifferentiableExpectedHypervolumedaulton2020} to optimize the samples from multiple nodes.

\subsubsection*{Computational overhead.}
The overhead of using structured DAG is high compared to the standard approaches.
During training and sampling, \BoGraph{} performs sequential operations without exploiting the independence between nodes to batch and distribute the workload.
Leveraging parallelization will reduce the computational time of \BoGraph{}.

\subsubsection*{Application domains.}
We have shown \BoGraph{} ability to find meaningful structure from logs and use that to optimize gem5-Aladdin.
Other computer systems will benefit from \BoGraph{} ability to make sense of logs in aiding the optimizer.
We are working on a case study on optimizing the latency of PostgreSQL \cite{PostgreSQLIntroductionConceptsmomjian2001}, \BoGraph{} is using the reported 370 metrics to find links between the metrics and the latency objective and tuning of thirty parameters.

\subsection{Limitations}
\subsubsection*{\BoGraph{}'s pipeline overhead}
\BoGraph{} maintains the system trace (previous evaluation, configurations, and logs) on disk, reading and writing these traces have both a memory and execution time overhead. This is a design choice we took for the following reasons:
First, the cost of the real system dominates the execution of the pipeline (\autoref{sc:exe_time}).
Secondly, this enables \BoGraph{} in the future to learn from distributed instances.
Finally, it provides a fault-tolerance mechanism for when the instance has to restart.

\textbf{Mitigation.} Every stage and component of the \BoGraph{} pipeline is configurable.
For example, the user can decide to reduce the frequency of a full causal structure discovery or include preprocessing steps that reduce the computation further by extending the preprocessing interface.
This flexibility offers some mitigation to the pipeline overhead.
Furthermore, to mitigate this overhead, \BoGraph{} can run a helper instance that runs the pipeline and communicate only the summarized information to the \BoGraph{} main loop.

\subsubsection*{Log annotation}
A key contribution of \BoGraph{} is providing a pipeline to make use of system logs and metrics to provide additional context that aids the system model.
Relying on the user to annotate their logs is one limitation of this work.
The annotation process involves providing a parser of the metrics or system logs; in gem5 and PostgresSQL experiments, it is a one-line regex expression.
Logs in computer systems follow a standard practice \cite{CharacterizingLoggingPracticesyuan2012} as they are often produced to be ingested into metric monitoring applications such as Prometheus \cite{PrometheusRunningInfrastructurebrazil2018}.
However, suppose the system is very different where the logs do not correlate with the system.
In that case, it is possible to disable the structure learning pipeline in \BoGraph{} and rely on a manual structure, resulting in a similar performance to the standard structure Bayesian Optimization method \cite{Dalibard2017}.
\section{Related work}
\textbf{Search-based auto-tuners.}
Reinforcement learning \cite{ReinforcementLearningIntroductionsutton1998}, random search \cite{RandomSearchHyperparameterbergstra2012},
evolutionary search (in PetaBricks) \cite{Ansel2010}, hill-climbing (in OpenTuner) \cite{Ansel2014}, or population-search \cite{PopulationBasedTrainingjaderberg2017} are simple auto-tuner to use and scale to many parameters.
However, they require many system evaluations, waste valuable expert knowledge, and are highly dependent on initializing seed (unstable) \cite{DeepReinforcementLearninghenderson2017}.
These drawbacks render them unsuitable for our problem due to Challenge \ref{challenge:expensive}.

\textbf{Dimensionality reduction.} OtterTune \cite{VanAken2017} used Factor Analysis \cite{EasyGuideFactorkline2014} to reduce the system dimensions.
However, as the OtterTune experiment expanded to a large system \cite{InquiryMachineLearningbasedvanaken2021},
it made apparent that computer system parameters have complex interactions that standard dimensionality reduction methods fail to capture (Challenge \ref{challenge:dependency}).
Principled Component Analysis \cite{PrincipalComponentAnalysiswold1987} suffer from a similar problem.

\textbf{BO surrogate models.}
RandomForest \cite{ClassificationRegressionRandomForestliaw2002} used in SMAC \cite{EvaluationSequentialModelbasedhutter2013} and HyperMapper \cite{PracticalDesignSpacenardi2019a} scales to many parameters.
Unfortunately, non-Bayesian methods underestimate the models' variance \cite{TakingHumanOutshahriari2015} meaning they explore less frequently and miss on finding the optimal configurations.
DeepGPs \cite{DeepGaussianProcessesdamianou2013} use several layers of GPs that reduce the objective dimensionality by finding a latent structure of it.
However, it requires many experiments to find a meaningful latent structure as it ignores existing expert knowledge.

\textbf{Parallel based auto-tuners.} AutoTVM \cite{LearningOptimizeTensorchen2018}, Hyperband \cite{HyperbandNovelBanditbasedli2017} and BOHB \cite{BOHBRobustEfficientfalkner2018}
are techniques that depend on the environment being cheap to terminate and easily parallelizable.
Computer systems are expensive and slow to evaluate; engineering them to launch many parallel instances is non-trivial without impacting the measurement sensitivity.

\textbf{Experts-driven methods.}
Ernest \cite{ErnestEfficientPerformancevenkataraman2016}, and RubberBand \cite{RubberBandCloudbasedHyperparametermisra2021} use hand-crafted models to measure the proposed configurations' effectiveness quickly leading to quick optimization.
While Causal Bayesian Optimization \cite{CausalBayesianOptimizationaglietti2020} assumes that a full causal graph of the system exists to speed up finding optimal system configurations.
\section{Conclusion}
\BoGraph{} is a framework for optimizing systems with many parameters and slow execution.
\BoGraph{} contextualizes the system through learning association between its metrics and parameters.
Furthermore, it provides an API to encode experts' knowledge as probabilistic models.
The contextualization leads to tuning more parameters in fewer experiments.
Using \BoGraph{} we optimized the design of an accelerator to improve its energy-latency objective by $2x$ in ${1}/{4}$ of the evaluations of the next best method.
For future work, we are applying \BoGraph{} to PostgreSQL, showing applicability to a range of computer systems.

\begin{acks}
  We would like to thank the reviewers for their valuable feedback.
  A special thanks to Thomas Vanderstichele, Mihai Bujanca, Yomna El-Serafy, Ibrahim Abdou, and Rana ElRashidy for their comments that improved the readability of the paper.
  This research was supported by the Alan Turing Institute.
\end{acks}

\FloatBarrier{}
\bibliography{references.bib}
\bibliographystyle{plain}
% \appendix
% \clearpage
% \import{sections/}{appendix}

\end{document}